\pdfoutput=1

\documentclass[11pt]{article}

\usepackage{acl}

\usepackage{times}
\usepackage{latexsym}

\usepackage[T1]{fontenc}

\usepackage[utf8]{inputenc}

\usepackage{microtype}

\title{2nd Swiss German Speech to Standard German Text Shared Task at SwissText 2022}

\author{Michel Pl{\"u}ss, Yanick Schraner, Christian Scheller, Manfred Vogel \\
  Institute for Data Science \\
  University of Applied Sciences and Arts Northwestern Switzerland \\
  Windisch, Switzerland \\
  \texttt{michel.pluess@fhnw.ch} \\}

\begin{document}
\maketitle
\begin{abstract}
We present the results and findings of the 2nd Swiss German speech to Standard German text shared task at SwissText 2022. Participants were asked to build a sentence-level Swiss German speech to Standard German text system specialized on the Grisons dialect. The objective was to maximize the BLEU score on a test set of Grisons speech. 3 teams participated, with the best-performing system achieving a BLEU score of 70.1.
\end{abstract}

\section{Introduction}

The topic of this task is automatic speech recognition (ASR) for Swiss German. Swiss German is a family of German dialects spoken in Switzerland, see \citet{pluess2021}. Swiss German ASR is concerned with the transcription of Swiss German Speech to Standard German text and can be viewed as a speech translation task with similar source and target languages, see \citet{pluess2021}.

This task has two predecessors. The 2020 task \citep{pluess2020} provided a 70-hours labeled training set of automatically aligned Swiss German speech (predominantly Bernese dialect) and Standard German text. The test set also comprised mostly Bernese speech. The winning contribution by \citet{buechi2020} achieved a word error rate (WER) of 40.3 \%. The 2021 task \citep{pluess2021} provided an improved and extended 293-hours version of the 2020 training set, as well as a 1208-hours unlabeled speech dataset (predominantly Zurich dialect). The test set covered a large part of the Swiss German dialect landscape. The winning contribution by \citet{arabskyy2021} achieved a BLEU score \citep{papineni2002} of 46.0.

The goal of this task is to build a system able to translate Swiss German speech to Standard German text and optimize it for the Grisons dialect. To enable this, we provide the Swiss German labeled datasets SDS-200 \citep{pluess2022} and SwissDial \citep{dogan2021}, both including a substantial amount of Grisons speech, as well as the Standard German, French, and Italian labeled datasets of Common Voice 9.0 \citep{ardila2020}.

\section{Task Description}

The goal of the task is to build a sentence-level Swiss German speech to Standard German
text system specialized on the Grisons dialect. The submission with the best BLEU score on a test set of Grisons dialect speakers wins. Participants were encouraged to explore suitable transfer learning and fine-tuning approaches based on the Swiss German, Standard German, French, and Italian data provided.

\subsection{Data}

We provide 5 different training datasets to participants, all of which are collections of sentence-level transcribed speech. SDS-200 \citep{pluess2022} is a Swiss German dataset with 200 hours of speech from all major Swiss German dialect regions, of which 6 hours are in Grisons dialect. SwissDial \citep{dogan2021} is a Swiss German dataset with 34 hours of speech from all major Swiss German dialect regions, of which 11 hours are in Grisons dialect. From version 9.0 of the Common Voice project \citep{ardila2020}, we provide 1166 hours of Standard German, 926 hours of French, and 340 hours of Italian, all of which are official languages of Switzerland.

The test set was collected in a similar fashion to SDS-200 \citep{pluess2022}. It consists of 5 hours of sentence-level transcribed Grisons speech by 11 speakers, of which 8 are female and 3 are male. The set is divided into two equally sized parts, a public part (score on this part was displayed in the public ranking while the task was running) and a private part (final ranking is based on this part, was not available while the task was running). Two thirds of the texts are from Swiss newspapers and one third is from the minutes of parliament debates in Aarau and Wettingen. Care was taken to avoid any overlap between the Swiss newspaper sentences in this test set and the ones in SDS-200 \citep{pluess2022}.

\subsection{Evaluation}

The submissions are evaluated using BLEU score \citep{papineni2002}. Our evaluation script, which uses the NLTK \citep{bird2009} BLEU implementation, is open-source\footnote{\url{https://github.com/i4Ds/swisstext-2022-swiss-german-shared-task}}. The private part of the test set is used for the final ranking.

The test set contains the characters a-z, {\"a}, {\"o}, {\"u}, 0-9, and spaces, and the participants' models should support exactly these. Punctuation and casing are ignored for the evaluation. Numbers are not used consistently in the test set, so sometimes they are written as digits and sometimes they are spelled out. We create a second reference by automatically spelling out all numbers and use both the original and this adjusted reference in the BLEU score calculation. Participants were advised to have their models always spell out numbers. All other characters are removed from the submission (see evaluation script for details). Participants were therefore advised to replace each additional character in their training set with a sensible replacement.

\section{Results}

\begin{table}
\begin{center}
\begin{tabular}{|c|l|c|}
\hline \textbf{Rank} & \textbf{Team} & \textbf{BLEU} \\ \hline
1 & Baseline & 70.1 \\
2 & Stucki et al. & 68.1 \\
3 & Nafisi et al. & 55.3 \\
\hline
\end{tabular}
\end{center}
\caption{\label{results-table} Final ranking of the shared task. The BLEU column shows the BLEU score on the private 50 \% of the test set. }
\end{table}

3 teams participated in the shared task, including our baseline. Table \ref{results-table} shows the final ranking.

Our baseline achieves a BLEU score of 70.1. We use the model \emph{Transformer Baseline} described in \citet{pluess2022}. We train the model from scratch on SDS-200, SwissDial, and the Standard German part of Common Voice. Contrary to \citet{pluess2022}, we employ a Transformer-based language model (LM) with 12 decoder layers, 16 attention heads, an embedding dimension of 512, and a fully connected layer with 1024 units. The LM is trained on 67M Standard German sentences. We use a beam width of 60 during decoding. The same model achieves 65.3 BLEU on the 2021 task test set \citep{pluess2021}.

Stucki et al. achieve a BLEU score of 68.1. They use an XLS-R 1B model \citep{babu2021}, pre-trained on 436K hours of unlabeled speech in 128 languages, not including Swiss German. They fine-tune the model on SDS-200 and SwissDial. A KenLM 5-gram LM \citep{heafield2011} trained on the German Wikipedia is employed.

Nafisi et al. achieve a BLEU score of 55.3. They use an XLS-R 1B model \citep{babu2021}, pre-trained on 436K hours of unlabeled speech in 128 languages, not including Swiss German. They fine-tune the model on SDS-200. No LM is employed.

\section{Conclusion}

We have described the 2nd Swiss German speech to Standard German text shared task at SwissText 2022. The best-performing system on the Grisons speech test set is our baseline with a BLEU score of 70.1. The same system achieves a BLEU score of 65.3 on the 2021 task test set \citep{pluess2021}, a relative improvement of 42 \% over the highest score of the 2021 task. This highlights the large progress in the field over the last year. The main drivers for this progress seem to be the new dataset SDS-200 \citep{pluess2022} as well as the use of models pre-trained on large amounts of unlabeled speech as demonstrated by the teams Stucki et al. and Nafisi et al., who employed XLS-R models \citep{babu2021}. The addition of an LM seems to be especially important for XLS-R models. The main difference between Nafisi et al. and Stucki et al. is that the latter add an LM, leading to a relative improvement of 23 \% BLEU.

On the other hand, none of the 3 participating teams made a significant effort to optimize their system for the Grisons dialect. The best approach to create an ASR system optimized for a specific dialect remains to be found in future work. Incorporating the provided French and Italian data for training is another possible direction for future research.

\bibliography{custom}
\bibliographystyle{acl_natbib}

\end{document}